# A Fast Keypoint Based Hybrid Method for Copy Move Forgery Detection

**Sunil Kumar[1], J. V. Desai[1] and Shaktidev Mukherjee[2]**

[1] *Mody University of Science and Technology, Lakshmangarh, India*
[2] *Moradabad Institute of Technology, Moradabad, India*

**Abstract:** Copy move forgery detection in digital images has become a very popular research topic in the area of image forensics. Due to the availability of sophisticated image editing tools and ever increasing hardware capabilities, it has become an easy task to manipulate the digital images. Passive forgery detection techniques are more relevant as they can be applied without the prior information about the image in question. Block based techniques are used to detect copy move forgery, but have limitations of large time complexity and sensitivity against affine operations like rotation and scaling. Keypoint based approaches are used to detect forgery in large images where the possibility of significant post processing operations like rotation and scaling is more. A hybrid approach is proposed using different methods for keypoint detection and description. Speeded Up Robust Features (SURF) are used to detect the keypoints in the image and Binary Robust Invariant Scalable Keypoints (BRISK) features are used to describe features at these keypoints. The proposed method has performed better than the existing forgery detection method using SURF significantly in terms of detection speed and is invariant to post processing operations like rotation and scaling. The proposed method is also invariant to other commonly applied post processing operations like adding Gaussian noise and JPEG compression.

## 1. INTRODUCTION

Digital images play a very significant role in the modern era. Almost every story published in newspapers and magazines uses digital images. Many times digital images are used to build some perception about celebrities and political leaders. In case of legal issues also, they are used as corroboratory evidences. So, before believing what we see, it is very much necessary to establish the truth of the image. Due to the growing influence of the digital images, the instances of manipulated images are also going up. Hence, the research area of image forgery detection has been growing very fast in recent times [1]. Broadly, the forgery detection techniques can be classified in to active and passive categories [2]. In case of active techniques, some prior information is embedded in the image at the time of capturing and later it can be used to verify the authenticity of the image. One such technique is watermarking [2]. In case of passive techniques, no prior information about the image in question is required. Copy move forgery or cloning, splicing and retouching are the main techniques of creating image forgery[2]. Retouching is performed by applying some filters to change the appearance of a person like the age or mood as shown in Figure 1. In splicing, some part of intended image is replaced by the content from some other image as shown in Figure 2. The statistical parameters of that region are quite different from the rest of the image. Hence, statistical techniques are applied to detect such forgeries. However, in copy move forgery, content from the same image is used to hide some region of the image [3] as shown in Figure 3. Copy move forgery is more prevalent and challenging due to its ease of performing and extent of manipulation. Also, to make it difficult for detection, the manipulator performs some additional operations like adding some noise, compressing the manipulated image, rotation and scaling of the duplicated region etc. The very fact about unedited captured images that no significant regions can be exactly same, is exploited in detecting copy move forgery. So, any duplication of a significant region is treated as a case of copy move forgery. Block based methods are used to divide the image into overlapping regions and then blocks are compared to find duplication.

*E-mail: skvasistha@acm.org, jagandesai@yahoo.com, mukherjee.shaktidev@gmail.com*

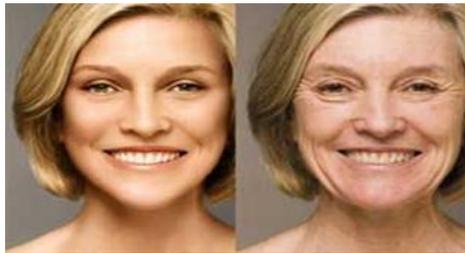

Figure 1. Original image on the left and retouched on the right

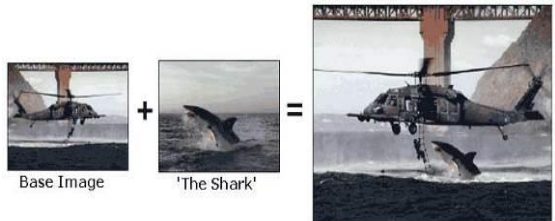

Figure 2. Image on the right is spliced using two images.

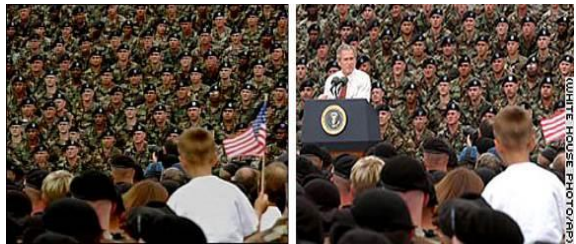

Figure 3. Copy move forgery. Image on the left is forged and original on the right

The main drawback of such methods is the large number of overlapping blocks in case of high resolution images. Also, these methods are not robust to large affine transformations applied to the duplicated region like rotation and scaling. A keypoint based hybrid method is proposed, which detects the keypoints, distinct in terms of intensity pattern around them and invariant to affine transforms like rotation and scaling. Speeded Up Robust Features (SURF) keypoints [4] are detected in the image and then Binary Robust Invariant Scalable Keypoints (BRISK) feature descriptors [5] are calculated at these keypoints and matched to find duplicated region. The time taken for detecting copy move forgery by the proposed method is compared with the existing keypoint based method using SURF and found significantly faster.

## 2. RELATED WORK

Most of the methods used to detect copy move forgery are based upon two approaches. One is the block based approach, in which the image in question is divided into overlapping blocks and these blocks acts as input to feature detection and matching phase. The other approach is the keypoint based approach, where keypoints are detected and descriptors at these keypoints are extracted and matched. The first block-based method was proposed by Fridrich et al. [6] based on discrete cosine transform (DCT). Popescu and Farid [7] altered the block representation and instead of DCT used principal component analysis (PCA). Sunil Kumar et al. [8] suggested a method by applying PCA on DCT domain to achieve robustness against both noise and JPEG compression. Huang et al [9] suggested an improved method using DCT coefficients. Luo et al [3] divided blocks into four sub-blocks, which were evaluated according to an average of red, blue and green color values. This method proved robust to attacks, such as JPEG compression, Gaussian blurring, and additive noise. An approach using combination of DWT and DCT is suggested in [10]. Discrete wavelet transform is used to reduce the size of image and then DCT is applied on low frequency component achieved by DWT. Singular value decomposition (SVD) is applied to each image block to yield a representation with reduced dimensions in [11]. Local binary patterns are used to get the binary feature vectors for robust and efficient matching in [12] [13]. Bayram et al. [14] applied Fourier Mellin Transform (FMT) and 1-D projection of log-polar values in a robust scheme for the detection of image forgeries. Local interest points have been widely used for image retrieval and object recognition, due to their robustness in dealing with numerous geometrical transformations (such as rotation and scaling) and occlusions. The method suggested in [15] use scale invariant feature transform [16] to locate the keypoints and match duplicated regions. Scale invariant feature transform is also used in [17], which is capable of detecting and describing clusters of points belonging to cloned regions. Amerini et al. [18] developed a SIFT-based method for the detection of copy move attacks and transformation recovery. Other variations of SIFT based methods are proposed in [19] [20]. Jaberi et al [21] used SIFT like feature MIFT to claim higher robustness. Another key point based method using speeded up robust features (SURF) is used in [22] and [23] which is faster than SIFT. This paper proposes a keypoint based method which employ both SURF and BRISK. Rigorous experiments have been conducted to demonstrate the speed improvement and robustness of the proposed method in dealing with post processing operations such as rotation, scaling, noise, and JPEG compression.

## 3. THE METHOD

The proposed method works as shown in Figure 4. The input image is converted to grayscale image. Keypoints are detected using SURF. To reduce the time taken for descriptor extraction and to make the matching process faster, BRISK descriptors are calculated at these keypoints. Hamming distance metric is used to match the binary features. Only nearest neighbors which are distinctively close are retained and others are discarded. Finally, the valid pair of keypoints are displayed on the image.

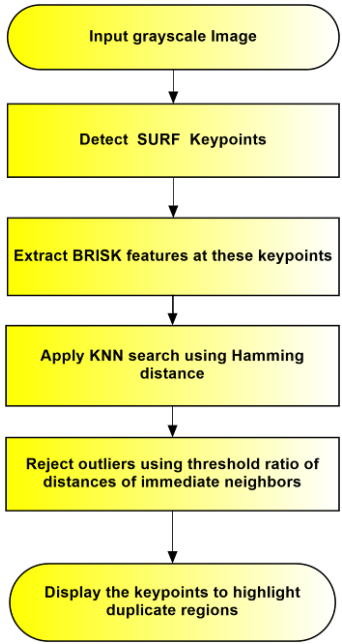

Figure 4. Proposed method framework

*A. Keypoint detection using SURF*

SURF was proposed for finding interest points or keypoints [4] and descriptor at these points in an image. The detector is based on Hessian matrix. The speed of the detector is due the use of integral image. Integral image $I_\Sigma$ of a given image $I$ at a given point $X=(x, y)$ is represented as the sum of all pixels in input image $I$ formed by the point X from the origin as in (1).

Integral image $I_\Sigma(X) = \sum_{i=0}^{i \leq x} \sum_{j=0}^{j \leq y} I(i,j)$  (1)

The SURF feature detector is based on the Hessian matrix because of its good performance in accuracy and speed. The Hessian matrix is defined as $H(x, \sigma)$ for a given point $X=(x, y)$ in an image $I$ as follows:

$$H(X, \sigma) = \begin{bmatrix} L_{xx}(X,\sigma) & L_{xy}(X,\sigma) \\ L_{xy}(X,\sigma) & L_{yy}(X,\sigma) \end{bmatrix}$$

Where $L_{xx}(X,\sigma)$ is convolution of the second order Gaussian derivative $\frac{\partial^2}{\partial x^2} g(\sigma)$ with image $I$ at $X$. $L_{xy}(X,\sigma)$, $L_{yy}(X,\sigma)$ are similarly defined and called as Laplacian of Gaussian. Gaussian kernel's standard deviation is $\sigma$. Working from this, the determinant of the Hessian for each pixel in the image is calculated and the values are used to find interest points. Box filters are used to approximate Gaussian second order derivatives. Using the integral image these approximate Gaussian second order derivatives can be evaluated very fast. Box size 9×9 and σ =1.2 represent the lowest scale (i.e. highest spatial resolution). The approximations are denoted by $D_{xx}$, $D_{yy}$ and $D_{xy}$. The determinant approximation of the Hessian matrix is calculated as in (2).

$det (H_{approx})= D_{xx}D_{yy} −(0.9D_{xy})$  (2)

Interest points are localized in scale and image space by applying non-maximum suppression in a 3 × 3 × 3 neighborhood. These maxima of the determinant of the approximated Hessian matrix are interpolated in scale and image space.

*B. BRISK descriptors*

Given a set of keypoints, the BRISK descriptor is composed as a binary string by concatenating the results of simple brightness comparison tests. In BRISK, the characteristic direction of each keypoint is identified to allow for orientation normalized descriptors to achieve rotation invariance which is important to general robustness. The key concept of the BRISK descriptor makes use of a pattern used for sampling the neighborhood of the keypoint. To negate aliasing effects the intensity values at these keypoints are smoothened using Gaussian. For a sampling point pair (X, Y) and smoothed intensity values at these points $I'(X)$ and $I'(Y)$ respectively, the local gradient $g(X, Y)$ is estimated as in (3).

$g(X,Y) = (X - Y) \frac{I'(X) - I'(Y)}{||X-Y||^2}$  (3)

Using two thresholds $\delta_{min}$ and $\delta_{max}$ the point pairs are divided into two sets S= {(X, Y) | ||X-Y|| < $\delta_{max}$ } and L= {(X,Y) | ||X-Y|| > $\delta_{min}$ }. Iterating through the point pairs in L overall characteristic pattern direction of a keypoint is calculated as in (4)

$g = \begin{pmatrix} g_x \\ g_y \end{pmatrix} = \frac{1}{|L|} \sum_{(X,Y) \in L} g(X,Y)$  (4)

Only long distance pairs are used for this computation. This is based on the assumption that local gradients annihilate each other and hence not necessary in the global gradient determination. Finally to get the rotation and scale normalized, BRISK descriptor sampling pattern is rotated by α=arctan2 ($g_y$, $g_x$) around the keypoint. The bit vector descriptor is assembled by performing all the short distance intensity comparisons of point pairs ($X^α$, $Y^α$) ∈ S (i.e. in the rotated pattern), such that each bit 'b' is defined as in (5).

$b = \begin{cases} 1 & if I'(Y^\alpha) > I'(X^\alpha) \\ 0 & otherwise \end{cases}$  (5)

$\forall$ $(X^\alpha, Y^\alpha) \in S$. The result is a bit vector of length 512 for the keypoints falling in the threshold range specified by $\delta_{min}$ and $\delta_{max}$ as shown in Figure 5.

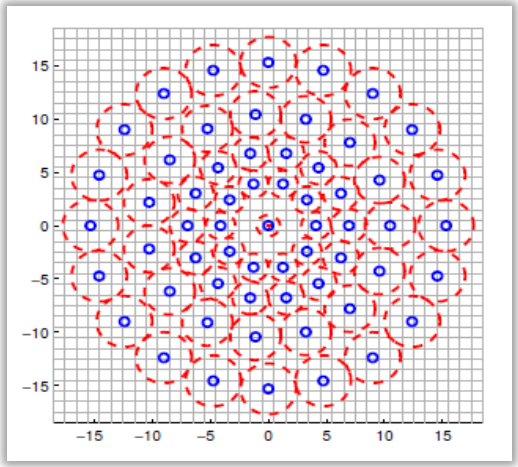

Figure 5: A typical BRISK feature vector extraction

### C. Feature matching

The binary features resulted from BRISK are matched for similarity using *knn* search. Hamming distance is used to find nearest neighbor. Hamming distance is the difference of the number of bits in two compared bit strings. All the nearest neighbors do not represent the legitimate duplicate keypoints. To discard the outliers distance ratio of the two immediate neighbors of a point is compared with a threshold value (ranges from 0.3-0.5). A keypoint is outlier if $(d_{ij}/d_{ik})>\rho$, where $d_{ij}$ and $d_{ik}$ are hamming distances of immediate neighbors of $i^{th}$ feature and $\rho$ is the threshold value.

## 4. EXPERIMENTAL RESULTS

### A. Experimental Setup

The method is tested on two databases [24] and [25]. Also some images are captured with personal camera. There are total 200 images taken for the experiment. The forged images have been prepared by applying copy move forgery with the post processing operations like rotation, scaling, noise addition and JPEG compression. The degree of rotation is varied upto 50 degrees. Scaling factor is varied from 1.1 to 2.0. Zero mean Gaussian noise is added with standard deviation from 0.02 to 0.1. Some images are JPEG compressed upto 40%. The algorithm is coded in MATLAB 2013a on a machine equipped with Intel i5 2.5 GHz processor with 8GB DDR3RAM.

### B. Performance evaluation and results

The method is tested using correct detection ratio which is ratio of valid keypoints to total matched keypoints without using morphological operations. It is compared with the existing method [23]. The proposed algorithm may be divided into three sections: (1) Extracting SURF keypoints; (2) define descriptors at these keypoints; (3) matching descriptors to locate the forged regions. The time taken by the first and third section is same for the proposed and the existing method are same as both extract the SURF keypoints and uses knn search.

TABLE I. COMPARISON OF DETECTION TIME WITH EXISTING SURF BASED METHOD

| Forged image name | Image size in pixels | Detection time (sec.) | |
|---|---|---|---|
| | | Existing method [23] | Proposed method |
| bullb_n_40.bmp | 480×640 | 0.56 | 0.18 |
| horses_copy_r8_gcs500.png | 1007×1520 | 0.8 | 0.54 |
| egyptian_copy_r2_gcs500.png | 1007×1520 | 1.07 | 0.80 |
| lone_cat_copy_r2_gcs500.jpg | 1007×1520 | 1.38 | 1.07 |
| writing_history_copy_r10_gcs500.png | 1007×1520 | 1.47 | 0.90 |
| hedge_copy_r10_gcs500.png | 1224×1632 | 0.61 | 0.31 |
| sails_copy_r2_gcs500.png | 1296×1944 | 1.73 | 1.3 |
| kore.bmp | 2592×3872 | 4.38 | 3.01 |
| maskcopy.bmp | 3039×2014 | 1.67 | 0.81 |
| mask_r35_s107.bmp | 3039×2014 | 1.72 | 1.36 |

TABLE II. PERFORMANCE OF THE PROPOSED METHOD IN TERMS OF KEYPOINT MATCHED

| Forged image name | Image size in pixels | Keypoints | | Correct detection ratio (%) | Threshold ($\rho$) |
|---|---|---|---|---|---|
| | | matched | valid | | |
| bullb_n_40.bmp | 480×640 | 54 | 54 | 100 | 0.9 |
| horses_copy_r8_gcs500.png | 1007×1520 | 160 | 154 | 96.25 | 0.3 |
| egyptian_copy_r2_gcs500.png | 1007×1520 | 274 | 272 | 99.27 | 0.3 |
| lone_cat_copy_r2_gcs500.jpg | 1007×1520 | 800 | 800 | 100 | 0.3 |
| writing_history_copy_r10_gcs500.png | 1007×1520 | 52 | 52 | 100 | 0.3 |
| hedge_copy_r10_gcs500.png | 1224×1632 | 96 | 92 | 95.83 | 0.4 |
| sails_copy_r2_gcs500.png | 1296×1944 | 124 | 115 | 92.74 | 0.3 |
| kore.bmp | 2592×3872 | 6358 | 6344 | 99.78 | 0.3 |
| maskcopy.bmp | 3039×2014 | 854 | 848 | 99.3 | 0.3 |
| mask_r35_s107.bmp | 3039×2014 | 368 | 343 | 93.21 | 0.4 |

However, the time taken by section 2 is different and comparison is provided in Table1 for descriptor extraction. The method is evaluated for stability by correct detection ratio, which is ratio of the keypoints in the forged region to the total keypoints detected. Another parameter termed relative detection efficiency is used to observe the behavior of the algorithm against the post processing operations of JPEG compression and noise addition. Relative detection efficiency is the ratio of correct keypoints detected in presence of post processing operation to the keypoints detected without using such operations.

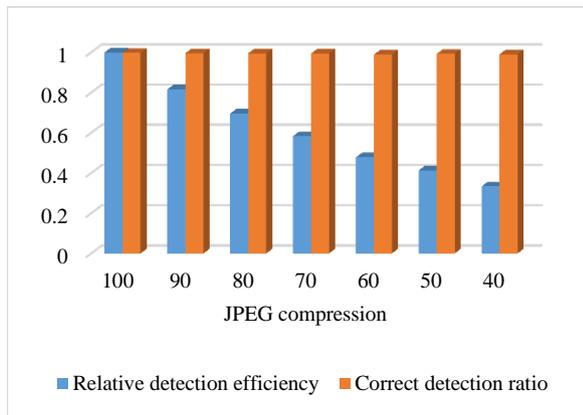

Figure 6: Performance against JPEG compression

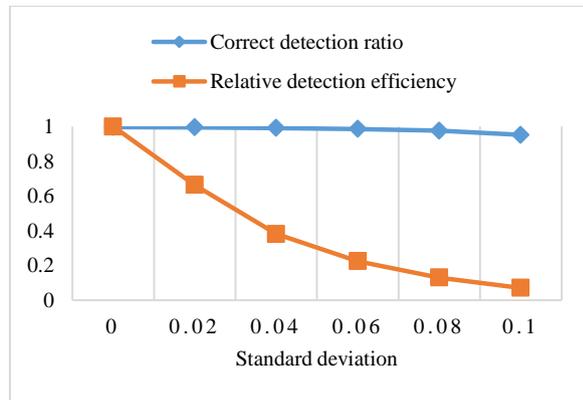

Figure 7: Performance against Gaussian noise

C. *Visual output of the proposed method*

The outputs for the images listed in the Table I are shown below:

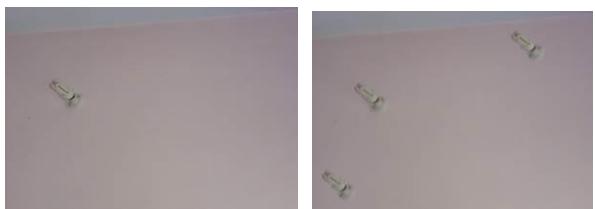

Figure 8: Original image 'bulb.png' (left) and forged image 'bullb_n_40.bmp' (right).

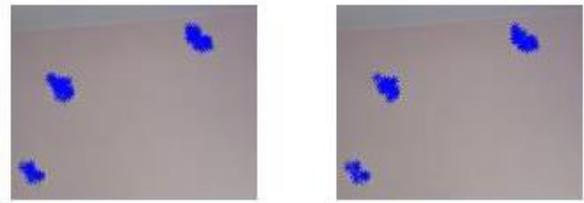

Figure 9: Detection results for the image in Figure 8 with all keypoints detected (left) and output image with valid matching keypoints (right).

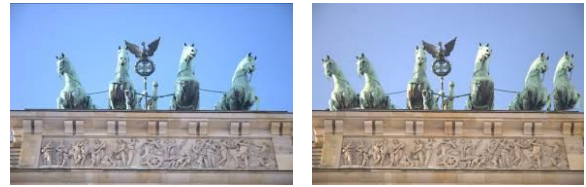

Figure 10. Original image'horses.png' (left) and forged image 'horses_copy_r8_gcs500.png' (right).

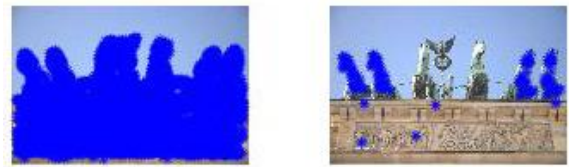

Figure 11: Detection results for the image in Figure 10 with all keypoints detected (left) and output image with valid matching keypoints (right).

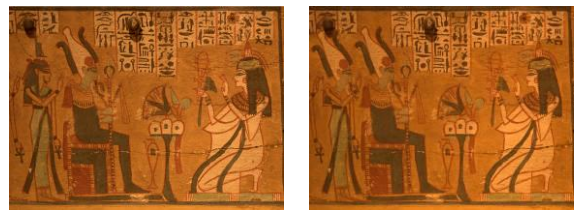

Figure 12: Original image' egyptian.png' (left) and forged image 'egyptian_copy_r2_gcs500.png' (right).

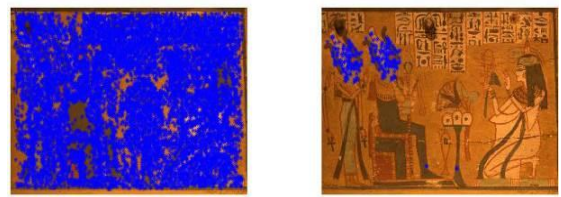

Figure 13: Detection results for the image in Figure 12 with all keypoints detected (left) and output image with valid matching keypoint (right).

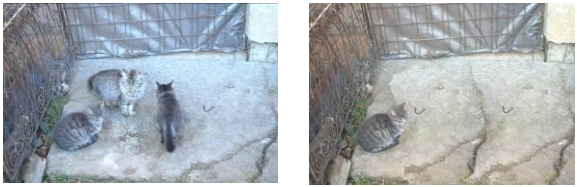

Figure 14: Original image 'lone_cat.png' (left) and forged image 'lone_cat_copy_r2_gcs500.jpg' (right).

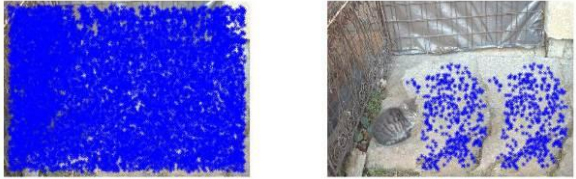

Figure 15: Detection results for the image in Figure 14 with all keypoints detected (left) and output image with valid matching keypoints (right).

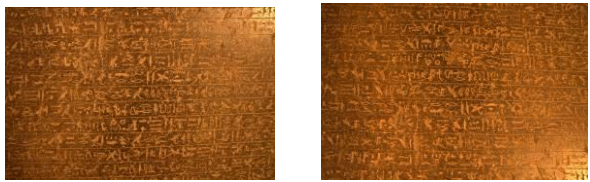

Figure 16. Original image 'writing_history.png' (left) and forged image 'writing_history_copy_r10_gcs500.png' (right).

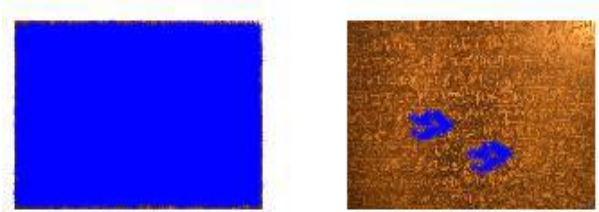

Figure 17: Detection results for the image in Figure 16 with all keypoints detected (left) and output image with valid matching keypoints (right).

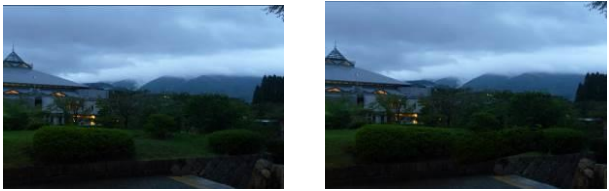

Figure 18. Original image 'hedge.png' (left) and forged image 'hedge_copy_r10_gcs500.png' (right).

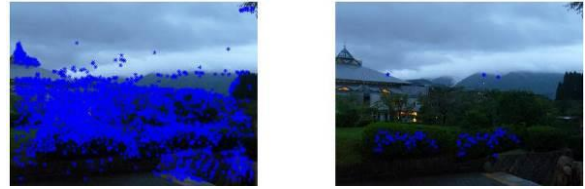

Figure 19. Detection results for the image in Figure 18 with all keypoints detected (left) and output image with valid matching keypoints (right).

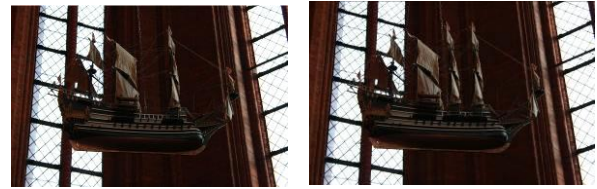

Figure 20. Original image 'sails.png' (left) and forged image 'sails_copy_r2_gcs500.png' (right).

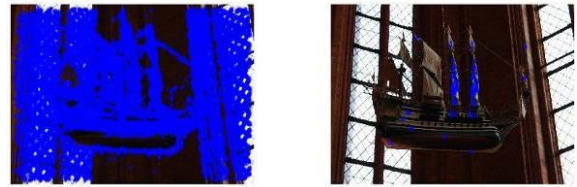

Figure 21. Detection results for the image in Figure 20 with all keypoints detected (left) and output image with valid matching keypoints (right).

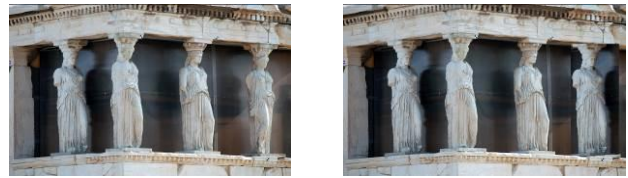

Figure 22: Original image 'kore.bmp (left) and forged image 'kore _copy.bmp' (right).

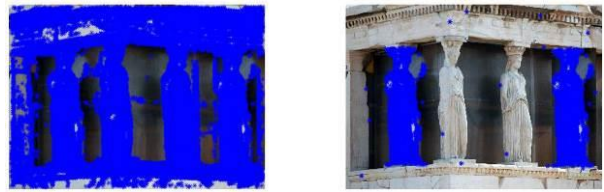

Figure 23: Detection results for the image in Figure 22 with all keypoints detected (left) and output image with valid matching keypoints (right)

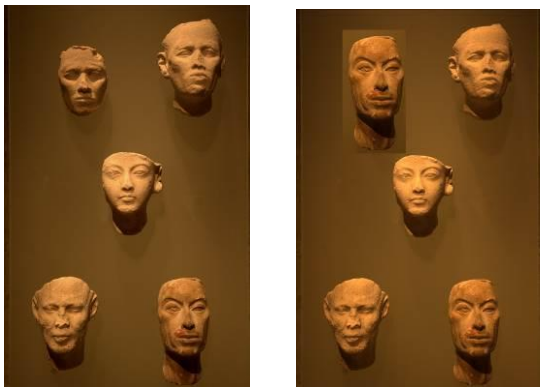

Figure 24: Original image 'mask.bmp' (left) and forged image 'maskcopy.bmp' (right)

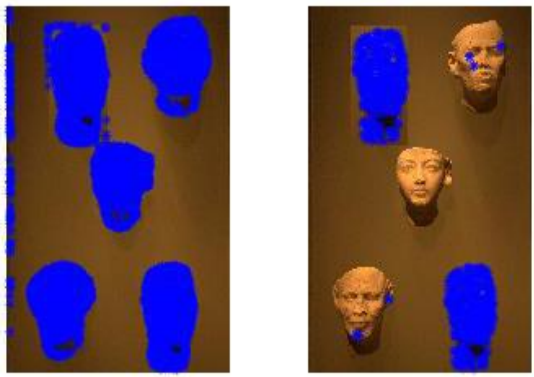

Figure 25: Detection results for the image in Figure 24 with all keypoints detected (left) and output image with valid matching keypoints (right).

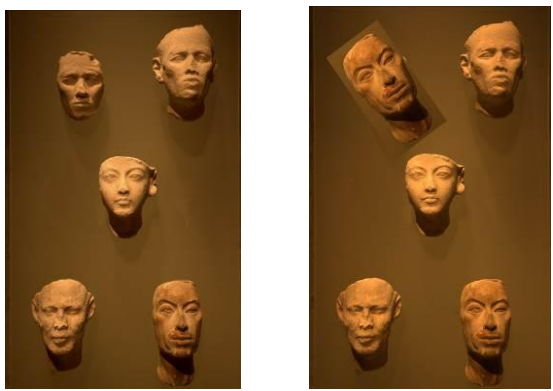

Figure 26: Original image 'mask.bmp' (left) and forged image 'mask_r35_s107.bmp' (right).

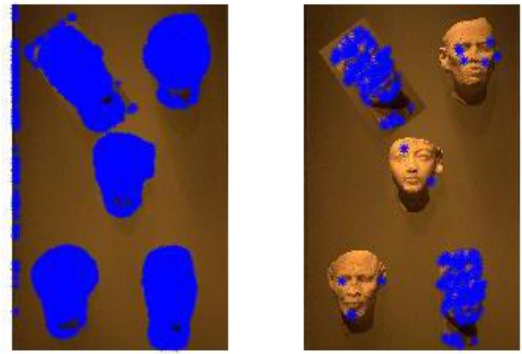

Figure 27: Detection results for the image in Figure 26 with all keypoints detected (left) and output image with valid matching keypoints (right)

*D. Discussion*

The experiment is conducted on image dataset with different size images. Figures 8-27 show detection results for different types of post processing operations including rotation and scaling. The time taken to detect forgery is not directly dependent upon the image size, rather it depends upon the duplicated region size and texture of the image. The number of keypoints generated and the corresponding descriptors processed are proportional to the duplicated region and brightness variations. As evident in case of 'maskcopy.bmp' and 'kore.bmp'. The size of the first one is greater than the later, but time taken as well as keypoints generated are more for the smaller image. Table II shows high stability of the method in terms of correct detection ratio. Figure 6 and Figure 7 show the performance of the method against JPEG compression and noise addition respectively. In case of small duplication regions and low contrast images the threshold ratio to identify good matches has to be quite high. The variation of threshold value (ρ) is tradeoff between number of valid keypoints and outliers. Low value of 'ρ' will cut the outliers but also restrict the number of valid keypoints.

## 5. CONCLUSION AND FUTURE SCOPE

The proposed method is significantly faster than the existing method using SURF. It is robust to affine transformations like rotation and scaling as well as other post processing operations like JPEG compression and Gaussian noise addition. In the matching process threshold is set manually depending upon the size of the copied area and the texture of the input image. So, there is further scope to devise the threshold automatically. For large values of rotation and scaling number of outliers also increases. Morphing techniques may be used to reduce the false keypoints and better localization of the duplicated region.

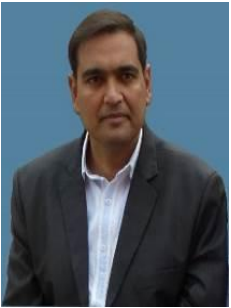

**Sunil Kumar** is working as Assistant Professor, Faculty of Engineering & Technology, Mody University of Science and Technology, Lakshmangarh. He has more than 12 years of UG and PG teaching experience after completing his M. Tech. in Computer Science & Engineering in 2002 from Kurukshetra University, Kurukshetra. He has supervised many B. Tech. and M. Tech projects. He has published many research papers in international conferences and journals. He is also reviewer of reputed journals. He is life member of Computer Society of India and current member of IEEE and ACM.

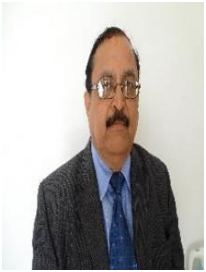

**Prof. J V. Desai** has completed his PhD form IIT Bombay. He has over 32 years' experience in teaching and research. He is currently Professor and Dean at Mody University of Science and Technology, Lakshmangarh. He has guided 7 Ph. Ds and many M. Tech. scholars. He has published numerous research articles in many international conferences and journals. He has research grants worth many lacs from government and private organizations. He is senior member of many professional bodies like IEEE. He is reviewer of many reputed international journals and conferences. His research interest are Soft Computing, Modeling & Simulation, Image & Signal Processing.

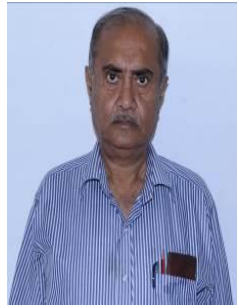

**Prof S. Mukherjee** graduated in Electrical Engineering from Patna University in 1968. After having Industrial experience for couple of years he did his Masters in Electrical Engineering from UOR in 1977 and since then is engaged in teaching in UOR and now IIT Roorkee. At present he is Professor and Director at Moradabad Institute of Technology, Moradabad and had been Vice Chancellor of Mody Institute of Technology and Science, Rajasthan from 2008 to 2010. He has supervised 7 Ph Ds and around 35 M. Tech dissertations during his teaching career so far. His areas of interests are system modeling, application of ANN in process control and other areas, order reduction of linear systems along with computer applications in Electrical Engineering like IEEE. He is reviewer of many reputed international journals and conferences.